\documentclass{article}

    \usepackage[final]{neurips_2025}

\usepackage[utf8]{inputenc} 
\usepackage[T1]{fontenc}    
\usepackage{hyperref}       
\usepackage{url}            
\usepackage{booktabs}       
\usepackage{amsfonts}       
\usepackage{nicefrac}       
\usepackage{microtype}      
\usepackage{xcolor}         
\usepackage{graphicx}
\usepackage{amsmath}
\usepackage{amssymb}
\usepackage{mathtools}
\usepackage{amsthm}
\usepackage{enumitem}
\usepackage{multirow}
\usepackage{multicol}

\usepackage{physics}
\usepackage{amsmath}
\usepackage{tikz}
\usepackage{mathdots}
\usepackage{yhmath}
\usepackage{cancel}
\usepackage{color}
\usepackage{siunitx}
\usepackage{array}
\usepackage{multirow}
\usepackage{amssymb}
\usepackage{gensymb}
\usepackage{tabularx}
\usepackage{extarrows}
\usepackage{booktabs}
\usetikzlibrary{fadings}
\usetikzlibrary{patterns}
\usetikzlibrary{shadows.blur}
\usetikzlibrary{shapes}
\usepackage{enumitem}

\title{Position: Untrained Machine Learning for Anomaly Detection by using 3D Point Cloud Data}

\author{%
  Juan Du \thanks{Corresponding author}\\
  Smart Manufacturing Thrust, Systems Hub \\
  The Hong Kong University of Science and Technology (Guangzhou) \\
  Guangzhou, China \\
  \texttt{juandu@ust.hk} \\
  \And
  Dongheng Chen \\
  Smart Manufacturing Thrust, Systems Hub \\
  The Hong Kong University of Science and Technology (Guangzhou) \\
  Guangzhou, China \\
  \texttt{chen\_dongheng@qq.com} 
}

\begin{document}
\maketitle
\begin{abstract}
  Anomaly detection based on 3D point cloud data is an important research problem and receives more and more attention recently. Untrained anomaly detection based on only one sample is an emerging research problem motivated by real manufacturing industries such as personalized manufacturing where only one sample can be collected without any additional labels and historical datasets. Identifying anomalies accurately based on one 3D point cloud sample is a critical challenge in both industrial applications and the field of machine learning. This paper aims to provide a formal definition of the untrained anomaly detection problem based on 3D point cloud data, discuss the differences between untrained anomaly detection and current unsupervised anomaly detection problems. Unlike trained unsupervised learning, untrained unsupervised learning does not rely on any data, including unlabeled data. Instead, they leverage prior knowledge about the surfaces and anomalies. 
 
We propose three complementary methodological frameworks: the Latent Variable Inference Framework that employs probabilistic modeling to distinguish anomalies; the Decomposition Framework that separates point clouds into reference, anomaly, and noise components through sparse learning; and the Local Geometry Framework that leverages neighborhood information for anomaly identification. Experimental results demonstrate that untrained methods achieve competitive detection performance while offering significant computational advantages, demonstrating up to a 15-fold increase in execution speed. The proposed methods provide viable solutions for scenarios with extreme data scarcity, addressing critical challenges in personalized manufacturing and healthcare applications where collecting multiple samples or historical data is infeasible.
  
\end{abstract}



\section{Introduction}
\label{Introduction}

Anomaly detection is a crucial research problem within the machine learning and quality control communities. Recently, there has been significant research focused on using machine learning for anomaly detection. Based on the availability of labels, anomaly detection methods can be categorized into three types: unsupervised (\cite{NEURIPS2024_99261adc}), supervised (\cite{NEURIPS2023_de670b9d}), and semi-supervised anomaly detection (\cite{NEURIPS2024_085b4b5d}). Unsupervised anomaly detection is particularly popular because it does not require labeled data, which is often scarce, allowing the algorithm to independently identify anomalies without the need for predefined labels. In contrast, supervised anomaly detection relies on labeled data, which limits the algorithm to detect only those anomalies that were encountered during training. Semi-supervised anomaly detection combines elements of both supervised and unsupervised approaches, enabling the handling of some labeled data alongside large amounts of unlabeled data. Among these three categories, unsupervised anomaly detection is particularly valuable because it eliminates the need for costly data labeling. Here in this paper, anomaly mainly refers as surface anomaly.

Based on the requirements for training data, unsupervised anomaly detection methods can be further divided into training-based methods and untrained methods. Training-based unsupervised methods rely on a sufficient number of anomaly-free training samples to learn the intrinsic patterns of the target object's surfaces, which then allows for the detection of anomalies. This type of unsupervised anomaly detection methods have proven effective in addressing data representation issues and managing the diverse nature of anomalies well.  These methods are equipped with sophisticated feature extractors that perform well, and they learn from anomaly-free samples without assuming the nature of potential anomalies, theoretically enabling them to detect any anomaly type. Nonetheless, in practical applications, anomalies with sparse characteristics or patterns resembling those of anomaly-free data can result in inaccurate detections. In addition, they are limited by their reliance on large amounts of anomaly-free training data, which can be a significant obstacle in scenarios where data samples are scarce such as personalized manufacturing and personalized healthcare.

In contrast, untrained methods do not require any anomaly-free samples for training, meaning that anomaly detection can be performed using just a single sample. Untrained unsupervised anomaly detection is becoming increasingly valuable as it addresses some of the limitations of traditional training-based unsupervised anomaly detection methods. While current untrained methods still require certain assumptions about anomalies, they offer a flexible and resource-efficient alternative, especially in situations where access to extensive anomaly-free datasets is not feasible such as personalized manufacturing and personalized healthcare. By not relying on predefined feature extractors or large datasets, untrained methods can adapt to various data and anomaly types. This flexibility allows them to effectively detect anomalies that might be rare, subtle, or closely resemble normal data. Such adaptability is crucial in real-world applications, where the nature of anomalies is often unpredictable and varied. Therefore, untrained unsupervised anomaly detection is essential for advancing the capability and reliability of anomaly detection systems.

Notably, untrained methods diverge fundamentally from the popular zero-shot and few-shot learning paradigms. While both untrained and zero-shot methods operate without task-specific training data, their fundamental approaches differ significantly. Zero-shot learning leverages pre-trained models on auxiliary datasets through semantic mappings to recognize unseen categories, making it suitable when prior defect templates are available but real anomaly samples are absent. Few-shot learning achieves rapid adaptation using minimal labeled samples, though its performance is strongly influenced by sample quality and quantity. In contrast, untrained methods bypass all training data entirely, achieving anomaly detection solely through geometric priors (e.g., surface smoothness, low-rankness) and physical constraints (e.g., sparsity assumptions). This distinction is particularly critical in personalized manufacturing contexts, where three critical constraints exist: (1) uniqueness of products eliminates the feasibility of obtaining labeled training data; (2) heterogeneous 3D scanners with varying point cloud densities preclude the use of pre-trained models; and (3) absence of CAD models for most customized designs invalidates template-based inspection methods. By directly analyzing geometric structural features from single samples and balancing computational efficiency with zero reliance on historical data, untrained methods maintain detection efficacy under extreme data scarcity, offering an innovative solution to industrial challenges of "zero sample availability" in personalized manufacturing and healthcare scenarios.

\color{black}

Developing untrained machine learning methods for anomaly detection by using 3D point cloud data presents three key challenges. First, effectively representing high-dimensional data for anomaly detection is a complex task. High-density 3D point cloud data can encompass millions of unstructured points, making it difficult to represent and creating hurdles for real-time computation. Second, having only one sample for anomaly detection makes it challenging to learn features and build an effective model. This necessitates a comprehensive utilization of prior knowledge, yet integrating this knowledge into the model remains a significant challenge. Finally, anomalies can appear locally, vary greatly, and are often sparse across objects. New anomalies can arise unexpectedly, and even anomalies of the same type can show significant differences, making detection difficult. Moreover, anomalies are typically sparse on object surfaces, occupying only a small part of the anomaly sample area, which makes the task of learning anomaly features even more difficult.

\begin{figure}
    \centering
    \includegraphics[width=1\linewidth]{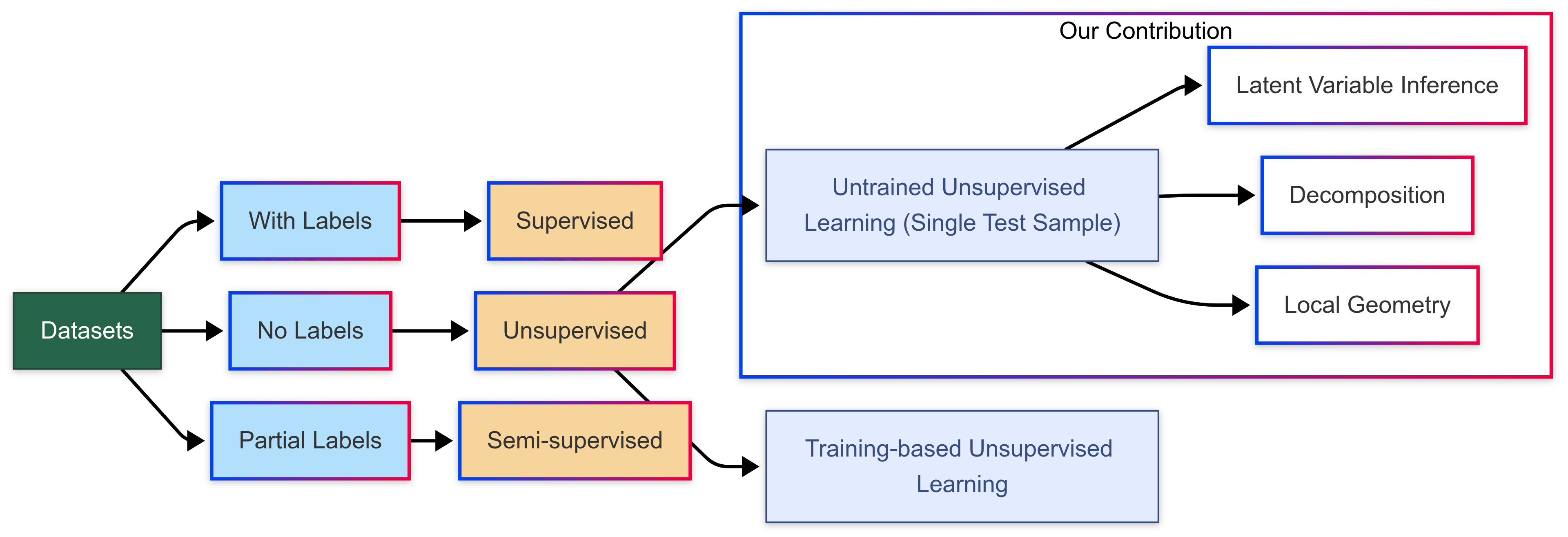}
    \caption{Overview and Our Contributions}
    \label{fig:enter-label}
\end{figure}

To address these challenges, several initial efforts have been made, including local geometry-based methods \cite{wang2023mvgcn},decomposition-based approaches \cite{tao2025pointsgrade}, and statistical latent variable inference methods \cite{tao2023anomaly}. A more detailed review will be provided in the following section. However, many research problems still need to be addressed. To advance the current state of untrained machine learning for anomaly detection, this paper first formally defines the anomaly detection problem via untrained machine learning, and three potential research methodology frameworks are proposed. Our contribution is illustrated in Figure \ref{fig:enter-label}. Three specific examples for each framework are also provided. Numerical studies are conducted to demonstrate the effectiveness and computational efficiency of untrained methods compared to existing training-based unsupervised learning methods. Finally, the paper concludes with a discussion and future outlook for untrained machine learning anomaly detection.

\section{Paper Review}

In the realm of manufacturing, anomaly detection using 3D point cloud data has witnessed a significant shift towards unsupervised methods, primarily due to the scarcity of annotated data. These unsupervised approaches can be broadly categorized into training-based methods and untrained methods, each with its own characteristics and applications.
Training-based unsupervised methods operate under the assumption that sufficient anomaly-free training samples are available. This enables them to learn the intrinsic patterns of normal surfaces, which are then used to detect anomalies. There are two main techniques within this category: feature embedding and reconstruction. Examples can be referred to  \cite{cao_complementary_2024,chu_shape-guided_2023,horwitz_back_2023,wang_multimodal_2023}.
Feature embedding-based methods involve a two-step process. First, latent features are extracted from anomaly-free training data. For example, in memory bank-based methods like Back To the Feature (BTF) \cite{horwitz_back_2023}, traditional descriptors such as FPFH are used to extract features from training point cloud patches, and these features are stored in a memory bank. The memory bank is then downsampled to represent the distribution of normal features. During inference, features that deviate significantly from this distribution are flagged as anomalies. Another approach is to use knowledge distillation (KD). Methods like 3D-ST \cite{bergmann_anomaly_2023} transfer knowledge from a pre-trained teacher network (e.g., RandLA-Net \cite{hu_learning_2022}) to a student network. The student network is trained on anomaly-free data to mimic the teacher's output, and discrepancies between the two are used to calculate anomaly scores during inference.

Reconstruction-based methods aim to achieve anomaly detection at the point level. Autoencoder-based methods, such as EasyNet \cite{chen_easynet_2023}, use a feature encoder and decoder to reconstruct  3D point cloud data. Trained only on anomaly-free data, they calculate anomaly scores based on the discrepancies between the input and the reconstructed data. For instance, EasyNet employs a multi-scale, multi-modality feature encoder-decoder for 3D depth map reconstruction, enabling real-time detection. However, some autoencoder-based methods like EasyNet and Cheating Depth \cite{zavrtanik_cheating_2024} require structured depth maps as input and struggle with unstructured point clouds. To address this, \cite{li_towards_2023} proposed the self-supervised Iterative Mask Reconstruction Network (IMRNet) for unstructured point cloud reconstruction and anomaly detection. Principal component analysis (PCA)-based methods, like those by \cite{von_enzberg_multiresolution_2016} and  \cite{zhang_automatic_2018}, identify normal patterns through principal components from training data and use them to reconstruct test samples for anomaly detection.
Despite their effectiveness, training-based methods are limited by the requirement for extensive anomaly-free training data. They also face challenges in accurately detecting anomalies with sparse properties or similar patterns to normal data. Feature embedding-based methods are good at localizing anomalies but have difficulty in obtaining accurate anomaly boundaries, while reconstruction-based methods can get more accurate anomaly boundaries but struggle with complex surface reconstruction, leading to higher false positive rates \cite{masuda_toward_2023,zavrtanik_cheating_2024,li_towards_2023,roth_towards_2022}.

Untrained methods, on the other hand, do not require training on a large dataset. They rely on prior knowledge to model the normal surface or possible anomalies (\cite{jovancevic_3d_2017}). Local geometry-based methods explore local geometric characteristics to detect anomalies. For example,  \cite{jovancevic_3d_2017} used a region-growing segmentation algorithm with local normal and curvature data to segment point cloud airplane surfaces. \cite{wei_microhardness_2021} developed local features to compute anomaly scores, and \cite{miao_pipeline_2022} used FPFH and normal vector aggregation for defect detection on gas turbine blades. Global geometry-based methods utilize the global shapes of manufacturing parts. Statistical-based methods, such as the one proposed by \cite{tao2023anomaly}, make assumptions about the shape of the product, like low-rankness and smoothness, and formulate the anomaly detection problem within a probabilistic framework. CAD model-based methods compare the point cloud with CAD models through rigid registration. \cite{zhao_defect_2023} used the standard iterative closest point (ICP) algorithm for defect detection of 3D printing products, and various improvements have been made to the ICP algorithm, such as the octree-based registration algorithm by \cite{he_octree-based_2023} to accelerate the process. Untrained methods can handle one single sample directly, which is an advantage in scenarios with limited data. 

In conclusion, unsupervised anomaly detection methods for 3D point cloud data in manufacturing have made significant progress. Training-based methods are suitable for scenarios with abundant anomaly-free data, while untrained methods offer a solution for situations where data is scarce. 
{The primary objective of this paper is to systematically define the untrained machine learning problem for 3D point cloud defect detection and highlight the irreplaceable significance of untrained methods in personalized manufacturing and other possible fields, aiming to draw greater academic attention to this field.}

\section{Problem Statement}
In this section, we provide a systematic mathematical formulation of an untrained anomaly detection problem based on 3D point cloud data. The definition of the problem is as follows.
Consider a 3D point cloud sample \( \mathbf{Y} = \{ \mathbf{Y}_i \in \mathbb{R}^3 \mid i = 1, 2, \ldots, N \} \), where \( N \) denotes the total number of points. The Anomaly Detection task aims to partition \( \mathbf{Y} \) into two disjoint subsets \( \mathbf{Y}_0 \) and \( \mathbf{Y}_1 \), satisfying:

\begin{enumerate}[itemsep=1pt]
    \item \( \mathbf{Y}_0 \cup \mathbf{Y}_1 = \mathbf{Y} \)
    \item \( \mathbf{Y}_0 \cap \mathbf{Y}_1 = \emptyset \)
    \item \( \mathbf{Y}_0 \) represents the reference surface points
    \item \( \mathbf{Y}_1 \) represents the anomaly points
\end{enumerate}
Furthermore, we make the following assumptions regarding the distribution of \( \mathbf{Y}_0 \) and \( \mathbf{Y}_1 \):
\begin{enumerate}[itemsep=1pt]
    \item \( \mathbf{Y}_0 \) is sufficiently dense, which is supported by contemporary 3D scanning technologies.
    \item \( \mathbf{Y}_1 \) exhibits sparse distribution characteristics.
\end{enumerate}
The point cloud anomaly detection problem can be formulated using the proposed 
 three complementary modeling frameworks: the Latent Variable Inference  Framework, the Decomposition Framework, and the Local Geometry Framework.

\subsection{Latent Variable Inference  Framework}
\label{sec:0}
From a latent variable inference perspective, the problem is formulated as a task defined by a mapping function $c : \mathbf{Y} \rightarrow \{0, 1\}$,

\begin{equation}
    c(\mathbf{Y}_i) = \begin{cases}
        0 & \text{if } \mathbf{Y}_i \in \mathbf{Y}_0 \text{ (reference surface point)} \\
        1 & \text{if } \mathbf{Y}_i \in \mathbf{Y}_1 \text{ (anomaly point)}
    \end{cases}
\end{equation}

Here $c(\mathbf{Y}_i)$ is a latent variable. The latent variable inference  problem is solved through a probabilistic approach. The relationships among variables \(\mathcal{C}\), \(\mathbf{Y}\), and the parameter set \(\mathbf{\Theta}\) can be characterized via the joint likelihood function \(p(\mathbf{Y}, \mathcal{C}|\mathbf{\Theta})\). Here, \(\mathbf{\Theta}\) encompasses parameters associated with the reference surface representation.

The optimal latent variable inference  set \(\mathcal{C}\) is then obtained by maximizing the joint likelihood:
\begin{equation}
    \mathcal{C} = \arg\max\limits_{\mathcal{C}}p(\mathbf{Y}, \mathcal{C}|\mathbf{\Theta})
\end{equation}

\begin{figure}
    \centering
    \includegraphics[width=0.8\linewidth]{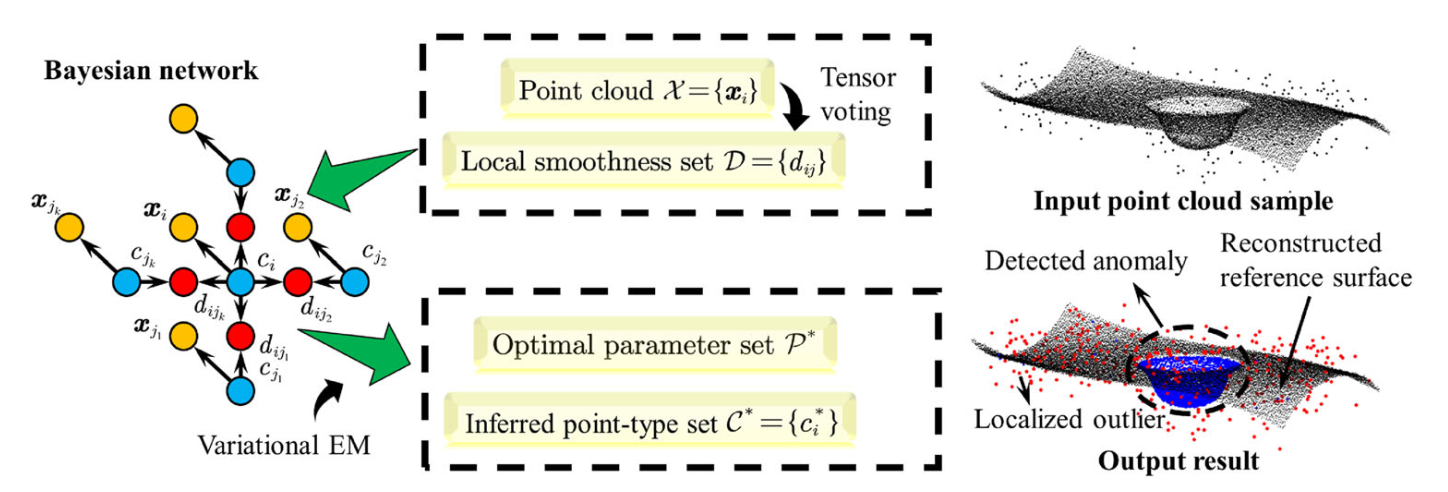}
    \caption{Example of Latent Variable Inference Framework \cite{tao_anomaly_2023}}
    \label{fig:1}
\end{figure}

An example of the Latent Variable Inference framework is provided by \cite{tao2023anomaly}, whom establishes probabilistic relationships between point locations and neighborhood smoothness, as shown in Figure \ref{fig:1}. This approach models the dependencies between surface points and their local geometric features, allowing for more precise anomaly identification in complex 3D structures. More details can be found in Appendix \ref{apLV}.

The Latent Variable Inference framework is suitable for the object surface where point relationship can be modeled. If it is hard to model the statistical relationship between points and point distributions, the following decomposition framework and local-geometry based framework can be considered. 
\color{black}
\subsubsection{Parameter Estimation and Latent Variable Inference}

To determine the optimal latent variable inference  set $\mathcal{C}$ by maximizing the joint likelihood function $p(\mathbf{Y}, \mathcal{C}|\mathbf{\Theta})$, we need to first estimate the parameter set $\mathbf{\Theta}$ that characterizes the reference surface representation. This involves a two-stage process: parameter estimation followed by latent variable inference.

\textbf{Parameter Estimation}

The parameter set $\mathbf{\Theta}$ must be inferred from the observed point cloud data through statistical methods. Given the available prior information about both anomaly and reference surfaces, we establish an appropriate probabilistic model. Since direct maximization of the joint likelihood is often intractable, we employ iterative approaches such as the Expectation-Maximization (EM) algorithm or its variants.

The EM algorithm alternates between:
\begin{itemize}
    \item \textbf{E-step}: Computing the expected value of the log-likelihood function with respect to the current estimate of parameters:
    \begin{equation}
        Q(\mathbf{\Theta}|\mathbf{\Theta}^{(t)}) = \mathbb{E}_{\mathcal{C}|\mathbf{Y},\mathbf{\Theta}^{(t)}}[\log p(\mathbf{Y}, \mathcal{C}|\mathbf{\Theta})]
    \end{equation}
    \item \textbf{M-step}: Updating the parameter estimates by maximizing this expected log-likelihood:
    \begin{equation}
        \mathbf{\Theta}^{(t+1)} = \arg\max\limits_{\mathbf{\Theta}} Q(\mathbf{\Theta}|\mathbf{\Theta}^{(t)})
    \end{equation}
\end{itemize}


\textbf{Latent Variable Inference}

Once the parameter set $\mathbf{\Theta}$ has been estimated, the optimal latent variable inference  set $\mathcal{C}$ can be determined using one of the following two techniques:

1. \textbf{Maximum a posteriori (MAP) estimation}:
\begin{equation}
    \mathcal{C}_{\text{MAP}} = \arg\max\limits_{\mathcal{C}} p(\mathcal{C}|\mathbf{Y},\mathbf{\Theta})
\end{equation}

2. \textbf{Maximum likelihood estimation (MLE)}:
\begin{equation}
    \mathcal{C}_{\text{MLE}} = \arg\max\limits_{\mathcal{C}} p(\mathbf{Y}|\mathcal{C},\mathbf{\Theta})p(\mathcal{C}|\mathbf{\Theta})
\end{equation}




The choice between MAP and MLE depends on the specific characteristics of the problem and available prior information. For problems with strong prior knowledge about the distribution of anomalies, MAP estimation may yield more robust results.

The mean field variational EM algorithm provides an efficient approach for estimating both the parameter set $\mathbf{\Theta}$ and the latent inference variables $\mathcal{C}$. This approach is particularly effective when dealing with large point clouds, as it balances computational efficiency with latent variable inference accuracy.

For surfaces where statistical point relationships are difficult to model, alternative approaches such as the decomposition framework or local geometry-based framework mentioned earlier may be more appropriate. These alternatives will be discussed in detail in subsequent sections.

\textbf{Parameter Tuning Considerations}

Following our parameter estimation and latent variable inference framework, appropriate parameter tuning is essential for optimal performance. The effectiveness of our statistical model depends significantly on the proper selection of the tuning parameter set. When domain knowledge is available, tuning parameters can be selected based on prior understanding of surface characteristics and anomaly patterns. This informed approach leverages existing expertise about the physical properties of the reference and anomaly surfaces. For scenarios lacking domain knowledge, we employ standard parameter tuning methods:
\begin{itemize}
    \item \textbf{Grid search}: Systematically evaluates model performance across predefined parameter spaces.
    \item \textbf{Analytical derivation}: For well-formulated optimization problems, maximum values of certain parameters can be mathematically derived.
    \item \textbf{Bisection method}: Efficiently searches the parameter space when parameters have monotonic effects on latent variable inference performance.
\end{itemize}

The parameter tuning process integrates seamlessly with our latent variable inference  framework, enhancing robustness across diverse surface types and point cloud characteristics. Experimental results demonstrating the effectiveness of these tuning strategies are presented in Section 5.
\color{black}

\subsection{Decomposition Framework}
\label{sec:1}
The data decomposition frameworks assume the point cloud \(\mathbf{Y}\) can be represented as a superposition of three components:
\begin{equation}
    \mathbf{Y} = \mathbf{X} + \mathbf{A} + \mathbf{E}
\end{equation}
where:
\begin{itemize}[itemsep=1pt]
    \item \(\mathbf{X} = [\mathbf{x}_1,\ldots,\mathbf{x}_N]^T \in \mathbb{R}^{N \times 3}\), denotes the reference surface component, representing the underlying regular geometric structure.
    \item \(\mathbf{A} = [\mathbf{a}_1,\ldots,\mathbf{a}_N]^T \in \mathbb{R}^{N \times 3}\), represents the anomaly component, capturing structural deviations from the reference surface.
    \item \(\mathbf{E} = [\mathbf{e}_1,\ldots,\mathbf{e}_N]^T \in \mathbb{R}^{N \times 3}\), accounts for measurement noise inherent in the data acquisition process.
\end{itemize}
    
As anomaly usually sparsely exists on the surfaces, we propose corresponding solution methods. For the decomposition-based formulation, we develop a sparse learning approach to recover the anomaly component as follows:
\begin{equation}
    \min_{\mathbf{A}} J(\mathbf{A}) = L(\mathbf{A}; \mathbf{X}, \mathbf{\Theta}) + \lambda p_s(\mathbf{A})
    \label{eq:optimization}
\end{equation}
where:
\begin{itemize}[itemsep=1pt]
    \item \(L(\mathbf{A}; \mathbf{X}, \mathbf{\Theta})\) is a loss function that quantifies the fitting residuals between \(\mathbf{Y}\) and \(\mathbf{X} + \mathbf{A}\), as well as enforces smoothness constraints on the reference surface parameters \(\mathbf{\Theta}\).
    \item \(p_s(\mathbf{A})\) is a penalty term promoting row-sparsity in \(\mathbf{A}\), and non-zero rows indicate the anomaly locations.
    \item \(\lambda\) is a tuning parameter controlling the sparsity level.
    \item \(\mathbf{\Theta}\) represents the estimated parameters of the reference surface.
\end{itemize}

\begin{figure}
    \centering
    \includegraphics[width=0.8\linewidth]{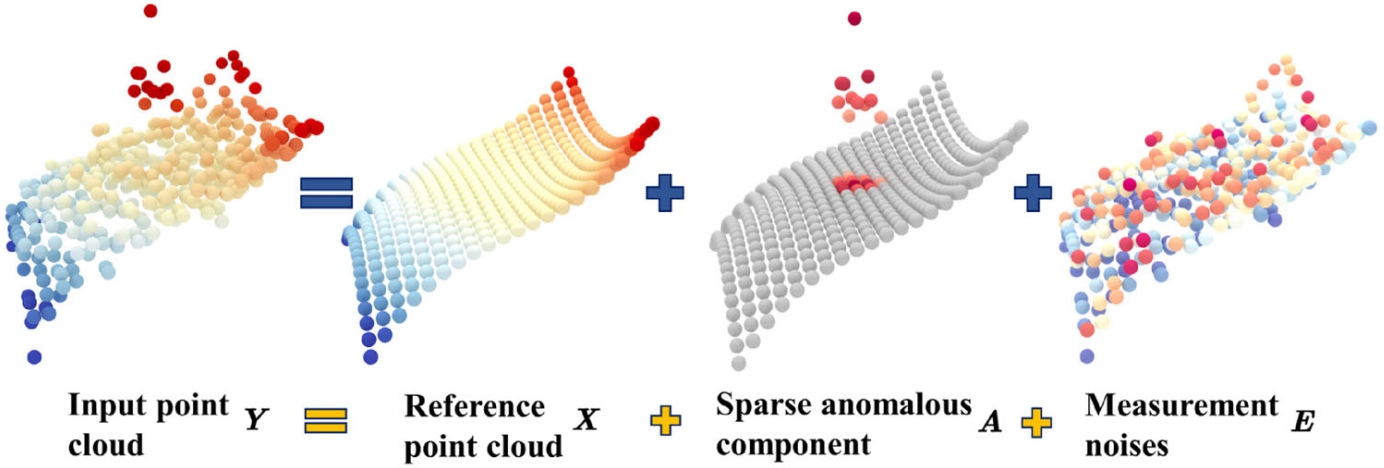}
    \caption{Example of Decomposition Framework \cite{tao2025pointsgrade}}
    \label{fig:2}
\end{figure}

An example of the Decomposition-based framework is provided by \cite{tao2025pointsgrade}, which achieves sparse anomaly recovery through graph-based smoothness constraints and group LOG penalties. This method effectively isolates anomalous components in the 3D point cloud data as shown in Figure \ref{fig:2}. The framework offers a robust strategy for identifying subtle surface deviations within large datasets. More details can be referred  to the Appendix \ref{ApD}.

Decomposition framework is useful for the surfaces where anomaly points can be separated from the normal reference surfaces. In addition, decomposition framework models the object surface as a whole, thereby enabling the accurate anomaly detection.  


\subsection{Local Geometry Framework}
\label{sec:2}
A classical perspective suggests that anomaly detection can be achieved according to neighborhood information. For any point, we can select its k nearest neighbors  and compute its Point Feature Histograms (PFH). These histograms possess invariance to rigid body transformations. After obtaining the Point Feature Histograms, we can employ machine learning methods to train a classifier that distinguishes between normal and anomalous points, thereby completing the task. 

There are various local-geometry based feature learning methods mentioned in \cite{Du202X}, and key ideas are utilizing local shape descriptors to extract the geometry feature from the neighbourhood. More details can be referred to \cite{Du202X}. Due to the local modeling of the object surfaces, local-geometry-based methods usually cannot detect anomaly boundaries very accurately, so the authors suggest prioritizing the first two frameworks. However, when the assumptions of the latent variable inference framework and decomposition framework fail, the local-geometry-based framework can serve as an alternative, which is more general in real applications.
\color{black}

\begin{figure}
    \centering
    \includegraphics[width=0.8\linewidth]{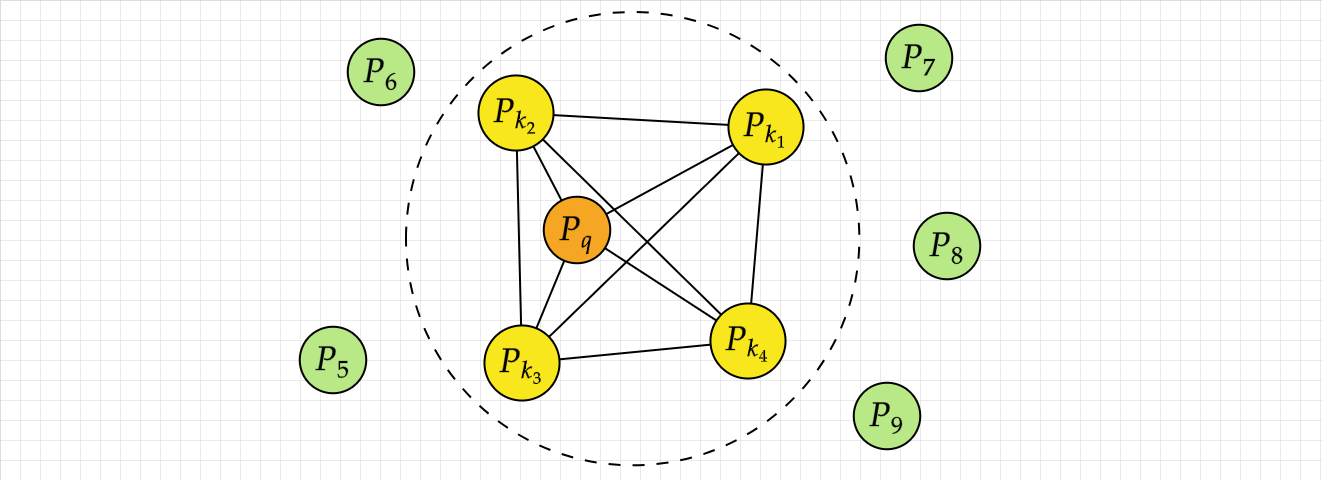}
    \caption{Example of Local-Geometry based Framework, the influence region diagram for a Point Feature Histogram. The query point (orange) and its k-neighbors (yellow) are fully interconnected in a dash circle. \cite{rusu_fast_2009}}
    \label{fig:3}
\end{figure}

An example of the Local Geometry Framework framework is provided by \cite{rusu2009fast}, which exemplifies local geometry analysis through Darboux frame construction and feature histogram computation, as ahown in Figure \ref{fig:3}. By analyzing point neighborhoods using invariant geometric features, the method provides a powerful tool for local anomaly detection in 3D point clouds. More details can be found in Appendix \ref{apL}.

\textbf{Integrating Local Geometry Framework with Machine Learning}

Although the Local Geometry Framework may not be the optimal approach for anomaly detection by using 3D point cloud data, it offers advantages in requiring fewer assumptions, making it more general and traditional. As noted in Section \ref{sec:0} and \ref{sec:1}, the Latent Variable Inference and Decomposition Frameworks typically achieve better detection accuracy and are recommended as primary approaches for real-world challenges.
However, the Local Geometry Framework can be enhanced through integration with machine learning models, which represents a promising direction for improvement. This integration must be tailored to specific applications, as demonstrated in  where tensor voting (a local shape descriptor) was successfully employed to measure surface smoothness.
The flexibility of local geometry methods combined with the pattern recognition capabilities of machine learning models offers a balanced approach that preserves generality while improving boundary detection accuracy. Such hybrid approaches may serve as effective alternatives when the assumptions required by the primary frameworks cannot be satisfied in particular application domains.

\begin{figure}
    \centering
    \includegraphics[width=0.8\linewidth]{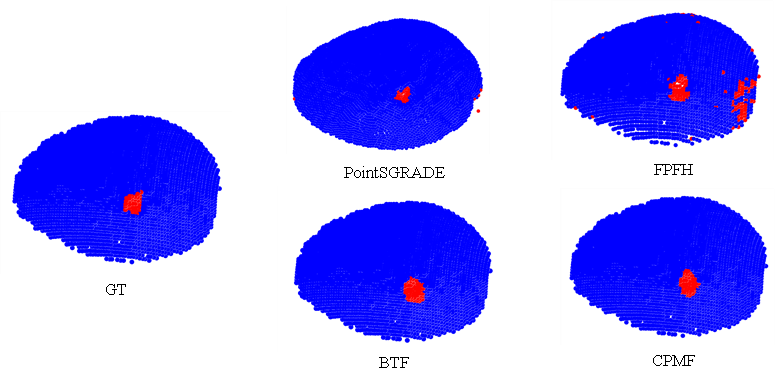}
    \caption{Anomaly detection results of different methods.}
    \label{fig:anomaly}
\end{figure}

\section{Experiments and Analysis}

\begin{table*}[htbp]
\centering
\caption{Comparative analysis of anomaly detection performance}
\label{table:detection_performance}
\begin{tabular}{llcccc}
\toprule
\multicolumn{2}{c}{Methods} & FOR$\downarrow$ & FPR$\downarrow$ & BA$\uparrow$ & DICE$\uparrow$ \\
\midrule
\multirow{2}{*}{Untrained methods} 
& PointSGRADE & 0.4208 & \textbf{0.0004} & 0.7893 & \textbf{0.7212} \\
& FPFH & \textbf{0.1463} & 0.0293 & \textbf{0.9122} & 0.4217 \\
\midrule
\multirow{2}{*}{Training-based methods} 
& BTF (CVPR23) & 0.2682 & 0.0104 & 0.8606 & 0.5825 \\
& CPMF (PR24) & \textbf{0.1951} & \textbf{0.0078} & 0.8985 & 0.6734 \\
\bottomrule
\end{tabular}
\end{table*}


\begin{table*}[htbp]
\centering
\caption{Computational efficiency comparison({1$\times$ A6000})}
\label{table:time_comparison}
\begin{tabular}{llcccc}
\toprule
\multicolumn{2}{c}{Methods} & Points & Training (s) & Inference (s) & Total (s) \\
\midrule
\multirow{2}{*}{Untrained methods}
& PointSGRADE & 3114 & - & 3.96 & 3.96 \\
& FPFH & 3114 & - & 0.626 & 0.626 \\
\midrule
\multirow{2}{*}{Training-based methods} 
& BTF (CVPR23) & 3114 & 15.39 & 0.36 & 15.75 \\
& CPMF (PR24) & 3114 & 94.99 & 0.76 & 95.75 \\
\bottomrule
\end{tabular}
\end{table*}

\textbf{Data Description and Experimental Setup:} Our evaluation uses the MVTec3D dataset containing 3D scans of industrial components with surface anomaly annotations. We standardized all point clouds to approximately 3,000 points through uniform sampling to ensure comparison fairness. The original dataset is divided into training (normal samples) and testing (containing both normal and anomalous samples) datasets following standard practice.

\textbf{Implementation Details:} 
\begin{itemize} 
\item \textbf{Untrained Methods:} Evaluated geometry-based approaches, including decomposition methods (PointSGRADE) and local descriptor methods (FPFH), using only testing data without training. 
\item \textbf{Training-based Methods:} Compared with memory bank approaches utilizing pre-trained features, specifically BTF (\cite{Horwitz_2023_CVPR}) and CPMF (\cite{Cao_2024}), the memory bank was constructed using 300 training samples with corresponding construction time shown in Table \ref{table:time_comparison}. 

\end{itemize}

\textbf{Evaluation Metrics} The evaluation metrics (pointwise) are defined as: $\text{FOR} = \frac{\text{FN}}{\text{FN} + \text{TN}}$, $\text{FPR} = \frac{\text{FP}}{\text{FP} + \text{TN}}$, $\text{BA} = \frac{1}{2}(\frac{\text{TP}}{\text{TP} + \text{FN}} + \frac{\text{TN}}{\text{TN} + \text{FP}})$, and $\text{DICE} = \frac{2\text{TP}}{2\text{TP} + \text{FP} + \text{FN}}$, where TP, TN, FP, and FN represent true positives, true negatives, false positives, and false negatives, and positive refers to a data point detected as anomalous. These evaluation indices are commonly used indices for anomaly detection by using 3D point cloud data (\cite{Du202X}).

\textbf{Key Findings:} Table 
\ref{table:detection_performance} reports detection metrics (FOR, FPR, BA, DICE) averaged across all testing points, while Table \ref{table:time_comparison} measures computational costs including memory bank construction time for training-based methods and inference time of a single sample. 

\begin{itemize} 
\item \textbf{Computational Efficiency:} Untrained methods demonstrate significant computational advantages, requiring no training overhead. FPFH achieves the fastest total execution time (0.626s), while PointSGRADE completes processing in 3.96s—both substantially faster than training-based methods which require 15.75s to 95.75s total time including training.
\item \textbf{Detection Performance:} Each evaluation metric reveals at least one untrained method achieving optimal performance. PointSGRADE excels in DICE score (0.7212) and FPR (0.0004), while FPFH achieves the best FOR (0.1463) and BA (0.9122), demonstrating that untrained methods can match or exceed training-based approaches across different performance criteria.
\end{itemize}

\textbf{Summary:} Untrained methods achieve competitive performance in 3D anomaly detection while requiring no training data, making them particularly valuable for industrial scenarios with insufficient data where conventional models fail to converge.

\section{Conclusion and Outlook}
This paper systematically defined untrained machine learning for anomaly detection by using 3D point cloud data, addressing critical challenges in scenarios without training samples. The significance of untrained methods lies in their ability to operate without pre-trained models or extensive anomaly-free datasets, making them particularly valuable for emerging applications such as personalized manufacturing and personalized healthcare where data scarcity is a fundamental constraint. Unlike traditional training-based unsupervised anomaly detection approaches that require substantial computational resources and unlabelled data, untrained methods achieve remarkable adaptability through the integration of geometric priors and sparse learning principles.

We formally establish three complementary frameworks to address different application scenarios. The \textit{Latent Variable Inference Framework} provides a probabilistic approach to distinguish anomalies from normal surfaces by introducing latent variables and corresponding statistical inference. The \textit{Decomposition Framework} separates point clouds into reference surface, anomaly, and noise components through sparse learning, enabling precise anomaly localization. The \textit{Local Geometry Framework} leverages neighborhood geometric features for anomaly identification when prior knowledge about surface structure is insufficient. 

As a position paper, our primary contribution lies in advocating for increased attention to untrained methodologies within the academic community. While significant advances have been achieved in training-based approaches, the applicability and irreplaceable value of untrained methods remain paramount. These methods provide both computational efficiency and superior task-specific performance for real-time applications in low-data scenarios. This work establishes foundational principles for robust anomaly detection in data-scarce environments and calls for future research to enhance computational efficiency for real-time applications and improve anomaly detection accuracy.

\bibliographystyle{unsrtnat}
\bibliography{example_paper} 

\begin{thebibliography}{31}
\providecommand{\natexlab}[1]{#1}
\providecommand{\url}[1]{\texttt{#1}}
\expandafter\ifx\csname urlstyle\endcsname\relax
  \providecommand{\doi}[1]{doi: #1}\else
  \providecommand{\doi}{doi: \begingroup \urlstyle{rm}\Url}\fi

\bibitem[Zhang et~al.(2024)Zhang, Li, Wu, Li, Lin, Hu, Li, and Jiang]{NEURIPS2024_99261adc}
Yu~Zhang, Ruoyu Li, Nengwu Wu, Qing Li, Xinhan Lin, Yang Hu, Tao Li, and Yong Jiang.
\newblock Dissect black box: Interpreting for rule-based explanations in unsupervised anomaly detection.
\newblock In A.~Globerson, L.~Mackey, D.~Belgrave, A.~Fan, U.~Paquet, J.~Tomczak, and C.~Zhang, editors, \emph{Advances in Neural Information Processing Systems}, volume~37, pages 84169--84196. Curran Associates, Inc., 2024.
\newblock URL \url{https://proceedings.neurips.cc/paper_files/paper/2024/file/99261adc8a6356b38bcf999bba9a26dc-Paper-Conference.pdf}.

\bibitem[Jiang et~al.(2023)Jiang, Hou, Zheng, Han, Huang, Wen, Hu, and Zhao]{NEURIPS2023_de670b9d}
Minqi Jiang, Chaochuan Hou, Ao~Zheng, Songqiao Han, Hailiang Huang, Qingsong Wen, Xiyang Hu, and Yue Zhao.
\newblock Adgym: Design choices for deep anomaly detection.
\newblock In A.~Oh, T.~Naumann, A.~Globerson, K.~Saenko, M.~Hardt, and S.~Levine, editors, \emph{Advances in Neural Information Processing Systems}, volume~36, pages 70179--70207. Curran Associates, Inc., 2023.
\newblock URL \url{https://proceedings.neurips.cc/paper_files/paper/2023/file/de670b9d118229d09d9a9bd9dec2598b-Paper-Datasets_and_Benchmarks.pdf}.

\bibitem[Qiao et~al.(2024)Qiao, Wen, Li, Lim, and Pang]{NEURIPS2024_085b4b5d}
Hezhe Qiao, Qingsong Wen, Xiaoli Li, Ee-Peng Lim, and Guansong Pang.
\newblock Generative semi-supervised graph anomaly detection.
\newblock In A.~Globerson, L.~Mackey, D.~Belgrave, A.~Fan, U.~Paquet, J.~Tomczak, and C.~Zhang, editors, \emph{Advances in Neural Information Processing Systems}, volume~37, pages 4660--4688. Curran Associates, Inc., 2024.
\newblock URL \url{https://proceedings.neurips.cc/paper_files/paper/2024/file/085b4b5d1f81ad9e057ad2b3de922ad4-Paper-Conference.pdf}.

\bibitem[Wang et~al.(2023{\natexlab{a}})Wang, Sun, Jin, Kong, and Yue]{wang2023mvgcn}
Yinan Wang, Wenbo Sun, Jionghua Jin, Zhenyu Kong, and Xiaowei Yue.
\newblock Mvgcn: Multi-view graph convolutional neural network for surface defect identification using three-dimensional point cloud.
\newblock \emph{Journal of Manufacturing Science and Engineering}, 145\penalty0 (3):\penalty0 031004, 2023{\natexlab{a}}.

\bibitem[Tao and Du(2025)]{tao2025pointsgrade}
Chengyu Tao and Juan Du.
\newblock Pointsgrade: Sparse learning with graph representation for anomaly detection by using unstructured 3d point cloud data.
\newblock \emph{IISE Transactions}, 57\penalty0 (2):\penalty0 131--144, 2025.

\bibitem[Tao et~al.(2023{\natexlab{a}})Tao, Du, and Chang]{tao2023anomaly}
Chengyu Tao, Juan Du, and Tzyy-Shuh Chang.
\newblock Anomaly detection for fabricated artifact by using unstructured 3d point cloud data.
\newblock \emph{IISE Transactions}, 55\penalty0 (11):\penalty0 1174--1186, 2023{\natexlab{a}}.

\bibitem[Cao et~al.(2024{\natexlab{a}})Cao, Xu, and Shen]{cao_complementary_2024}
Yunkang Cao, Xiaohao Xu, and Weiming Shen.
\newblock Complementary pseudo multimodal feature for point cloud anomaly detection.
\newblock \emph{Pattern Recognition}, 156:\penalty0 110761, December 2024{\natexlab{a}}.
\newblock ISSN 00313203.

\bibitem[Chu et~al.(2023)Chu, Liu, Hsieh, Chen, and Liu]{chu_shape-guided_2023}
Yu-Min Chu, Chieh Liu, Ting-I Hsieh, Hwann-Tzong Chen, and Tyng-Luh Liu.
\newblock Shape-{Guided} {Dual}-{Memory} {Learning} for {3D} {Anomaly} {Detection}.
\newblock \emph{Proceedings of the 40 th International Conference on Machine Learning}, 2023.

\bibitem[Horwitz and Hoshen(2023{\natexlab{a}})]{horwitz_back_2023}
Eliahu Horwitz and Yedid Hoshen.
\newblock Back to the {Feature}: {Classical} {3D} {Features} are ({Almost}) {All} {You} {Need} for {3D} {Anomaly} {Detection}.
\newblock In \emph{2023 {IEEE}/{CVF} {Conference} on {Computer} {Vision} and {Pattern} {Recognition} {Workshops} ({CVPRW})}, Vancouver, BC, Canada, June 2023{\natexlab{a}}. IEEE.
\newblock ISBN 979-8-3503-0249-3.

\bibitem[Wang et~al.(2023{\natexlab{b}})Wang, Peng, Zhang, Yi, Wang, and Wang]{wang_multimodal_2023}
Yue Wang, Jinlong Peng, Jiangning Zhang, Ran Yi, Yabiao Wang, and Chengjie Wang.
\newblock Multimodal {Industrial} {Anomaly} {Detection} via {Hybrid} {Fusion}.
\newblock In \emph{2023 {IEEE}/{CVF} {Conference} on {Computer} {Vision} and {Pattern} {Recognition} ({CVPR})}, pages 8032--8041, Vancouver, BC, Canada, June 2023{\natexlab{b}}. IEEE.
\newblock ISBN 979-8-3503-0129-8.

\bibitem[Bergmann and Sattlegger(2023)]{bergmann_anomaly_2023}
Paul Bergmann and David Sattlegger.
\newblock Anomaly {Detection} in {3D} {Point} {Clouds} using {Deep} {Geometric} {Descriptors}.
\newblock In \emph{2023 {IEEE}/{CVF} {Winter} {Conference} on {Applications} of {Computer} {Vision} ({WACV})}, pages 2612--2622, Waikoloa, HI, USA, January 2023. IEEE.
\newblock ISBN 978-1-6654-9346-8.

\bibitem[Hu et~al.(2022)Hu, Yang, Xie, Rosa, Guo, Wang, Trigoni, and Markham]{hu_learning_2022}
Qingyong Hu, Bo~Yang, Linhai Xie, Stefano Rosa, Yulan Guo, Zhihua Wang, Niki Trigoni, and Andrew Markham.
\newblock Learning {Semantic} {Segmentation} of {Large}-{Scale} {Point} {Clouds} {With} {Random} {Sampling}.
\newblock \emph{IEEE Transactions on Pattern Analysis and Machine Intelligence}, 44\penalty0 (11):\penalty0 8338--8354, November 2022.
\newblock ISSN 1939-3539.

\bibitem[Chen et~al.(2023)Chen, Xie, Liu, Wang, Luo, Wang, and Zheng]{chen_easynet_2023}
Ruitao Chen, Guoyang Xie, Jiaqi Liu, Jinbao Wang, Ziqi Luo, Jinfan Wang, and Feng Zheng.
\newblock {EasyNet}: {An} {Easy} {Network} for {3D} {Industrial} {Anomaly} {Detection}.
\newblock In \emph{Proceedings of the 31st {ACM} {International} {Conference} on {Multimedia}}, pages 7038--7046, Ottawa ON Canada, October 2023. ACM.
\newblock ISBN 979-8-4007-0108-5.

\bibitem[Zavrtanik et~al.(2024)Zavrtanik, Kristan, and Skoaj]{zavrtanik_cheating_2024}
Vitjan Zavrtanik, Matej Kristan, and Danijel Skoaj.
\newblock Cheating {Depth}: {Enhancing} {3D} {Surface} {Anomaly} {Detection} via {Depth} {Simulation}.
\newblock In \emph{2024 {IEEE}/{CVF} {Winter} {Conference} on {Applications} of {Computer} {Vision} ({WACV})}, pages 2153--2161, Waikoloa, HI, USA, January 2024. IEEE.
\newblock ISBN 979-8-3503-1892-0.

\bibitem[Li et~al.(2023)Li, Xu, Gu, Zheng, Gao, and Wu]{li_towards_2023}
Wenqiao Li, Xiaohao Xu, Yao Gu, Bozhong Zheng, Shenghua Gao, and Yingna Wu.
\newblock Towards {Scalable} {3D} {Anomaly} {Detection} and {Localization}: {A} {Benchmark} via {3D} {Anomaly} {Synthesis} and {A} {Self}-{Supervised} {Learning} {Network}, November 2023.

\bibitem[Von~Enzberg and Al-Hamadi(2016)]{von_enzberg_multiresolution_2016}
Sebastian Von~Enzberg and Ayoub Al-Hamadi.
\newblock A {Multiresolution} {Approach} to {Model}-{Based} 3-{D} {Surface} {Quality} {Inspection}.
\newblock \emph{IEEE Transactions on Industrial Informatics}, 12\penalty0 (4):\penalty0 1498--1507, August 2016.
\newblock ISSN 1551-3203, 1941-0050.

\bibitem[Zhang et~al.(2018)Zhang, Zou, Lin, Xu, He, Gui, and Li]{zhang_automatic_2018}
Dejin Zhang, Qin Zou, Hong Lin, Xin Xu, Li~He, Rong Gui, and Qingquan Li.
\newblock Automatic pavement defect detection using {3D} laser profiling technology.
\newblock \emph{Automation in Construction}, 96:\penalty0 350--365, December 2018.
\newblock ISSN 09265805.

\bibitem[Masuda et~al.(2023)Masuda, Hachiuma, Fujii, Saito, and Sekikawa]{masuda_toward_2023}
Mana Masuda, Ryo Hachiuma, Ryo Fujii, Hideo Saito, and Yusuke Sekikawa.
\newblock Toward {Unsupervised} {3D} {Point} {Cloud} {Anomaly} {Detection} using {Variational} {Autoencoder}, April 2023.

\bibitem[Roth et~al.(2022)Roth, Pemula, Zepeda, Scholkopf, Brox, and Gehler]{roth_towards_2022}
Karsten Roth, Latha Pemula, Joaquin Zepeda, Bernhard Scholkopf, Thomas Brox, and Peter Gehler.
\newblock Towards {Total} {Recall} in {Industrial} {Anomaly} {Detection}.
\newblock In \emph{2022 {IEEE}/{CVF} {Conference} on {Computer} {Vision} and {Pattern} {Recognition} ({CVPR})}, pages 14298--14308, New Orleans, LA, USA, June 2022. IEEE.
\newblock ISBN 978-1-6654-6946-3.

\bibitem[Jovanevi et~al.(2017)Jovanevi, Pham, Orteu, Gilblas, Harvent, Maurice, and Brthes]{jovancevic_3d_2017}
Igor Jovanevi, Huy-Hieu Pham, Jean-Jos Orteu, Rmi Gilblas, Jacques Harvent, Xavier Maurice, and Ludovic Brthes.
\newblock 3d point cloud analysis for detection and characterization of defects on airplane exterior surface.
\newblock \emph{Journal of Nondestructive Evaluation}, 36:\penalty0 1--17, 2017.

\bibitem[Wei et~al.(2021)Wei, Chengyao, Shilin, and Kailin]{wei_microhardness_2021}
Shi Wei, Shi Chengyao, Chen Shilin, and Zhang Kailin.
\newblock A microhardness indentation point cloud segmentation method based on voxel cloud connectivity segmentation.
\newblock \emph{Measurement: Sensors}, 18:\penalty0 100124, 2021.

\bibitem[Miao et~al.(2022)Miao, Fu, Wu, Hao, Li, Hao, and Zhou]{miao_pipeline_2022}
Yubin Miao, Ruochong Fu, Hang Wu, Mengxiang Hao, Gang Li, Jiarui Hao, and Dengji Zhou.
\newblock Pipeline of turbine blade defect detection based on local geometric pattern analysis.
\newblock \emph{Engineering Failure Analysis}, 133:\penalty0 105965, 2022.

\bibitem[Zhao et~al.(2023)Zhao, Li, Xiao, and He]{zhao_defect_2023}
Xinyue Zhao, Quanzhi Li, Menghan Xiao, and Zaixing He.
\newblock Defect detection of 3d printing surface based on geometric local domain features.
\newblock \emph{The International Journal of Advanced Manufacturing Technology}, 125\penalty0 (1):\penalty0 183--194, 2023.
\newblock Number: 1 Publisher: Springer.

\bibitem[He et~al.(2023)He, Ma, Li, Hao, Wang, and Wang]{he_octree-based_2023}
Yan He, Wen Ma, Yufeng Li, Chuanpeng Hao, Yulin Wang, and Yan Wang.
\newblock An octree-based two-step method of surface defects detection for remanufacture.
\newblock \emph{International Journal of Precision Engineering and Manufacturing-Green Technology}, 10\penalty0 (2):\penalty0 311--326, 2023.
\newblock Number: 2 Publisher: Springer.

\bibitem[Tao et~al.(2023{\natexlab{b}})Tao, Du, and Chang]{tao_anomaly_2023}
Chengyu Tao, Juan Du, and Tzyy-Shuh Chang.
\newblock Anomaly detection for fabricated artifact by using unstructured {3D} point cloud data.
\newblock \emph{IISE Transactions}, 55\penalty0 (11):\penalty0 1174--1186, November 2023{\natexlab{b}}.
\newblock ISSN 2472-5854, 2472-5862.

\bibitem[Du et~al.(2025)Du, Tao, Cao, and Tsung]{Du202X}
J.~Du, C.~Tao, X.~Cao, and F.~Tsung.
\newblock 3d vision-based anomaly detection in manufacturing: A survey.
\newblock \emph{Frontiers of Engineering Management}, 2025.

\bibitem[Rusu et~al.(2009-05)Rusu, Blodow, and Beetz]{rusu_fast_2009}
Radu~Bogdan Rusu, Nico Blodow, and Michael Beetz.
\newblock Fast point feature histograms ({FPFH}) for 3d registration.
\newblock In \emph{2009 {IEEE} International Conference on Robotics and Automation}, pages 3212--3217. {IEEE}, 2009-05.
\newblock ISBN 978-1-4244-2788-8.

\bibitem[Rusu et~al.(2009)Rusu, Blodow, and Beetz]{rusu2009fast}
Radu~Bogdan Rusu, Nico Blodow, and Michael Beetz.
\newblock Fast point feature histograms (fpfh) for 3d registration.
\newblock In \emph{2009 IEEE international conference on robotics and automation}, pages 3212--3217. IEEE, 2009.

\bibitem[Horwitz and Hoshen(2023{\natexlab{b}})]{Horwitz_2023_CVPR}
Eliahu Horwitz and Yedid Hoshen.
\newblock Back to the feature: Classical 3d features are (almost) all you need for 3d anomaly detection.
\newblock In \emph{Proceedings of the IEEE/CVF Conference on Computer Vision and Pattern Recognition (CVPR) Workshops}, pages 2968--2977, June 2023{\natexlab{b}}.

\bibitem[Cao et~al.(2024{\natexlab{b}})Cao, Xu, and Shen]{Cao_2024}
Yunkang Cao, Xiaohao Xu, and Weiming Shen.
\newblock Complementary pseudo multimodal feature for point cloud anomaly detection.
\newblock \emph{Pattern Recognition}, 156:\penalty0 110761, December 2024{\natexlab{b}}.
\newblock ISSN 0031-3203.
\newblock \doi{10.1016/j.patcog.2024.110761}.
\newblock URL \url{http://dx.doi.org/10.1016/j.patcog.2024.110761}.

\bibitem[Schwartz et~al.(2024)Schwartz, Arbelle, Karlinsky, Harary, Scheidegger, Doveh, and Giryes]{schwartz2024maedaymaezeroshot}
Eli Schwartz, Assaf Arbelle, Leonid Karlinsky, Sivan Harary, Florian Scheidegger, Sivan Doveh, and Raja Giryes.
\newblock Maeday: Mae for few and zero shot anomaly-detection, 2024.
\newblock URL \url{https://arxiv.org/abs/2211.14307}.

\end{thebibliography}

\appendix

\newpage

\section{Examples}

\subsection{Example of Latent Variable Inference Framework}
\label{apLV}
The Bayesian network structure presented in \citet{tao2023anomaly} establishes a probabilistic framework for anomaly detection in unstructured 3D point cloud data. This review examines its key components and theoretical foundations. {For an overview of the model architecture, refer to Figure \ref{fig:1}.}

\textbf{Core Components and Relationships}

The network models three fundamental relationships:

1) Point location ($x_i$) depends on point type ($c_i$):
\begin{equation}
p(x_i | C, X, D) = p(x_i | c_i)
\end{equation}

2) Local smoothness ($d_{ij}$) depends on adjacent point types:
\begin{equation}
d_{ij} = \|K_i-K_j\|_F^2
\end{equation}
where $K_i$ represents the structure-aware tensor of point $i$.

3) Point-type inference relies on spatial location and neighborhood smoothness:
\begin{equation}
p(c_i | C, X, D) = p(c_i | x_i, \{d_{ij}, c_j, j \in \mathcal{N}_i\})
\end{equation}

\textbf{Joint Distribution}

The network enables factorization of the joint distribution:
\begin{equation}
p(X, D, C) = \prod_i p(x_i | c_i)p(c_i) \prod_{j \in \mathcal{N}_i} p(d_{ij} | c_i, c_j)
\end{equation}
The structure's validity is established through the Markov property, ensuring that each node's distribution depends only on its Markov blanket. This property validates the conditional independence assumptions and enables efficient probabilistic inference.

To implement efficient inference in this probabilistic framework, we employ a mean field variational EM algorithm, which approximates the posterior distribution $p(C|X,D)$ with a factorized form $q(C) = \prod_i q(c_i)$. This approach iteratively optimizes the variational lower bound on the log-likelihood, enabling effective estimation of both latent variables $C$ and model parameters while maintaining computational tractability for large point clouds.

\subsection{Example of Decomposition Framework }
\label{ApD}
{Recent studies have proposed various decomposition-based methods for anomaly detection. As shown in Figure \ref{fig:2}, \citet{tao2025pointsgrade} introduced a novel framework with the loss function defined as:}
\begin{equation}
\min_{\mathbf{A}}J(\mathbf{A})=L(\mathbf{A};\mathbf{H},\mathbf{Y})+\lambda p_{s}(\mathbf{A})
\end{equation}
In this formulation, the loss function $L(\mathbf{A};\mathbf{H},\mathbf{Y})$ is formulated to quantify the fitting residuals between the observed point cloud $\mathbf{Y}$ and the sum of the reference point cloud $\mathbf{X}$ and the anomalous component $\mathbf{A}$. This quantification effectively measures how closely the combination of $\mathbf{X}$ and $\mathbf{A}$ can approximate $\mathbf{Y}$. Moreover, it enforces the smoothness of $\mathbf{X}$ through the graph-based smoothness metric $\mathbf{H}$.

The matrix $\mathbf{H}$ is derived from Graph Signal Processing (GSP) theories. By representing the reference point cloud $\mathbf{X}$ as a graph, $\mathbf{H}$ is constructed in such a way that $\mathbf{HX}$ can capture the smoothness of $\mathbf{X}$. Geometrically, the $i^{\text{th}}$ row of $\mathbf{HX}$, denoted as $(\mathbf{HX})_i$, can be interpreted as the difference between the $i^{\text{th}}$ point and the convex combination of its neighbors. This interpretation approximates the deviation of the $i^{\text{th}}$ point from the local plane. For a smooth $\mathbf{X}$, the value of $\mathbf{HX}$ is approximately zero. This property is integrated into the loss function to guarantee the smoothness of the reference surface.

The penalty term $p_{s}(\mathbf{A})$ is introduced with the aim of promoting the row-sparsity of $\mathbf{A}$. Given that anomalies are typically sparse, a row-sparse $A$ implies that only a small number of points are identified as anomalies. The authors selected the group LOG penalty, defined as $p_{s}(\mathbf{A})=\sum_{i}\log(\sqrt{\left\|a_{i}\right\|_{2}^{2}+\varepsilon}+\left\|a_{i}\right\|_{2})$, for this purpose, where the group LOG penalty can penalize different predictors more evenly. This characteristic enables a better recovery of the sparse anomalous component $\mathbf{A}$.

\begin{figure}[htbp]
    \centering
    \begin{minipage}{\textwidth}
        \centering
        \includegraphics[width=0.8\linewidth]{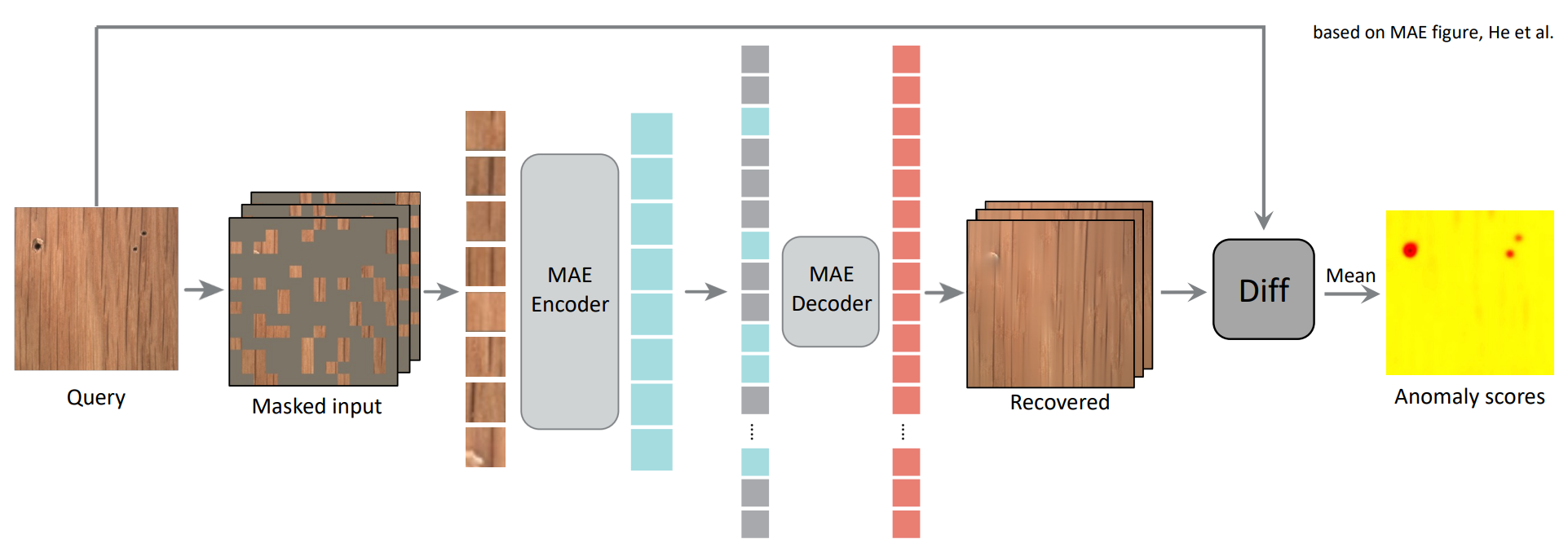}
        \caption{Example of Zero-Shot Framework\citet{schwartz2024maedaymaezeroshot}}
        \label{fig:4}
    \end{minipage}
    
    \vspace{1cm}
\end{figure}

\subsection{Example of Local-Geometry based Framework}
\label{apL}
{This example is sourced from the work presented in \citet{rusu2009fast}, whose computational framework is illustrated in Figure \ref{fig:3}. The approach unfolds through the following key steps:}

\textbf{Neighborhood Selection}
Given a point $p$ in a 3D point cloud, we first select all its neighbors enclosed within a sphere of radius $r$. Let $k$ denote the number of these neighbors, and we refer to this set as the $k$-neighborhood of $p$.

\textbf{Darboux Frame Construction and Feature Calculation}
For each pair of points $p_i$ and $p_j$ ($i \neq j$) in the $k$-neighborhood of $p$ with their estimated normals $n_i$ and $n_j$ (where $p_i$ is the point with a smaller angle between its associated normal and the line connecting the points), we define a Darboux $uvn$ frame:
\begin{align}
\mathbf{u} &= \mathbf{n}_i \\
\mathbf{v} &= (\mathbf{p}_j - \mathbf{p}_i) \times \mathbf{u} \\
\mathbf{w} &= \mathbf{u} \times \mathbf{v}\\
\alpha &= \mathbf{v} \cdot \mathbf{n}_j \\
\phi &= \frac{\mathbf{u} \cdot (\mathbf{p}_j - \mathbf{p}_i)}{\|\mathbf{p}_j - \mathbf{p}_i\|} \\
\theta &= \arctan(\mathbf{w} \cdot \mathbf{n}_j, \mathbf{u} \cdot \mathbf{n}_j)
\end{align}
Previous research sometimes included a fourth feature—the Euclidean distance from $p_i$ to $p_j$. However, recent experiments have demonstrated that its omission does not significantly impact robustness in certain cases, particularly in 2.5D datasets.

\textbf{Histogram Construction}
After computing these features for all point pairs in the $k$-neighborhood of $p$, we quantize these feature values and bin them into a histogram. The resulting histogram constitutes the Point Feature Histogram (PFH) at point $p$, characterizing the local geometric properties in its vicinity.

\textbf{Simplified Point Feature Histogram (SPFH) Calculation}
For each query point $p$, we initially compute only the relationships between itself and its neighbors, termed the Simplified Point Feature Histogram (SPFH).

\textbf{FPFH Computation from SPFH}
Subsequently, for each point $p$, we re-determine its $k$ neighbors. The Fast Point Feature Histogram (FPFH) at point $p$ is computed as:

\begin{equation}
FPFH(p) = SPF(p) + \frac{1}{k}\sum_{i=1}^{k}\frac{1}{\omega_k} \cdot SPF(p_k)
\end{equation}

\noindent
where $\omega_k$ represents the distance between query point $p$ and a neighbor point $p_k$ in a given metric space. This formulation combines the SPFH of point $p$ with a weighted sum of its neighbors' SPFHs, achieving reduced computational complexity compared to PFH while maintaining most of its discriminative power.

\section{Code}
This is the open-source code for our experiment (anonymized for privacy). You can access it through the following link:

\url{https://anonymous.4open.science/r/pointSGRADE-78E4/README.md}

\newpage
\section*{NeurIPS Paper Checklist}

The checklist is designed to encourage best practices for responsible machine learning research, addressing issues of reproducibility, transparency, research ethics, and societal impact. Do not remove the checklist: {\bf The papers not including the checklist will be desk rejected.} The checklist should follow the references and follow the (optional) supplemental material.  The checklist does NOT count towards the page
limit. 

Please read the checklist guidelines carefully for information on how to answer these questions. For each question in the checklist:
\begin{itemize}
    \item You should answer \answerYes{}, \answerNo{}, or \answerNA{}.
    \item \answerNA{} means either that the question is Not Applicable for that particular paper or the relevant information is Not Available.
    \item Please provide a short (1–2 sentence) justification right after your answer (even for NA). 
\end{itemize}

{\bf The checklist answers are an integral part of your paper submission.} They are visible to the reviewers, area chairs, senior area chairs, and ethics reviewers. You will be asked to also include it (after eventual revisions) with the final version of your paper, and its final version will be published with the paper.

The reviewers of your paper will be asked to use the checklist as one of the factors in their evaluation. While "\answerYes{}" is generally preferable to "\answerNo{}", it is perfectly acceptable to answer "\answerNo{}" provided a proper justification is given (e.g., "error bars are not reported because it would be too computationally expensive" or "we were unable to find the license for the dataset we used"). In general, answering "\answerNo{}" or "\answerNA{}" is not grounds for rejection. While the questions are phrased in a binary way, we acknowledge that the true answer is often more nuanced, so please just use your best judgment and write a justification to elaborate. All supporting evidence can appear either in the main paper or the supplemental material, provided in appendix. If you answer \answerYes{} to a question, in the justification please point to the section(s) where related material for the question can be found.

IMPORTANT, please:
\begin{itemize}
    \item {\bf Delete this instruction block, but keep the section heading ``NeurIPS Paper Checklist"},
    \item  {\bf Keep the checklist subsection headings, questions/answers and guidelines below.}
    \item {\bf Do not modify the questions and only use the provided macros for your answers}.
\end{itemize}


\section*{NeurIPS Paper Checklist}

\begin{enumerate}

\item {\bf Claims}
    \item[] Question: Do the main claims made in the abstract and introduction accurately reflect the paper's contributions and scope?
    \item[] Answer: \answerYes{}
    \item[] Justification: The paper claims to systematically define untrained machine learning for anomaly detection, propose three frameworks (Classification, Decomposition, and Local Geometry), and highlight their significance in personalized manufacturing and healthcare. These claims are accurately reflected in the detailed methodology sections (3.1-3.3) and experimental results (Section 4) that demonstrate computational efficiency and detection performance.

\item {\bf Limitations}
    \item[] Question: Does the paper discuss the limitations of the work performed by the authors?
    \item[] Answer: \answerYes{}
    \item[] Justification: Section 5.1 explicitly addresses limitations of the current methods, including PointSGRADE's strong smoothness assumptions that can increase false positive rates on complex surfaces, and the limited capacity of traditional geometric descriptors for complex surface representations. Section 5 further discusses these limitations and proposes future research directions.

\item {\bf Theory assumptions and proofs}
    \item[] Question: For each theoretical result, does the paper provide the full set of assumptions and a complete (and correct) proof?
    \item[] Answer: \answerYes{}
    \item[] Justification: The paper provides comprehensive mathematical formulations in Section 3, including assumptions about reference surfaces (Section 3, points 1-5) and theoretical derivations for the Classification Framework (3.1), Decomposition Framework (3.2), and Local Geometry Framework (3.3). Section 3.1 includes detailed parameter estimation processes using Expectation-Maximization algorithms and variational approaches.

\item {\bf Experimental result reproducibility}
    \item[] Question: Does the paper fully disclose all the information needed to reproduce the main experimental results of the paper to the extent that it affects the main claims and/or conclusions of the paper (regardless of whether the code and data are provided or not)?
    \item[] Answer: \answerYes{}
    \item[] Justification: Section 4 provides detailed experimental setup, including comparison methods (untrained vs. training-based approaches), point cloud sampling (approximately 3,000 points), and evaluation metrics (FOR, FPR, BA, DICE). Tables 1 and 2 present quantitative results that enable reproducibility, while Figure 1 shows visualization results.

\item {\bf Open access to data and code}
    \item[] Question: Does the paper provide open access to the data and code, with sufficient instructions to faithfully reproduce the main experimental results, as described in supplemental material?
    \item[] Answer: \answerNo{}
    \item[] Justification: While the paper references existing methods like PointSGRADE, FPFH, BTF, and CPMF in the experimental section, it does not explicitly mention providing open access to implementation code or datasets. 
    The dataset is a open dataset and the code will be released upon acceptance.

\item {\bf Experimental setting/details}
    \item[] Question: Does the paper specify all the training and test details (e.g., data splits, hyperparameters, how they were chosen, type of optimizer, etc.) necessary to understand the results?
    \item[] Answer: \answerYes{}
    \item[] Justification: Yes, the paper provides comprehensive experimental details including dataset specification (MVTec3D with 300 training samples), data preprocessing (uniform sampling to ~3,000 points), standard train/test splits following dataset conventions, implementation details for both untrained methods (PointSGRADE, FPFH) and training-based methods (BTF, CPMF), computational resources (1×A6000 GPU), and clearly defined evaluation metrics (FOR, FPR, BA, DICE) with their mathematical formulations.

\item {\bf Experiment statistical significance}
    \item[] Question: Does the paper report error bars suitably and correctly defined or other appropriate information about the statistical significance of the experiments?
    \item[] Answer: \answerNA{}
    \item[] Justification: Not applicable. As a position paper, we focus on comparing two distinct paradigms - untrained unsupervised learning versus training-based unsupervised learning - rather than conducting statistical hypothesis testing that would require error bars or significance measures.

\item {\bf Experiments compute resources}
    \item[] Question: For each experiment, does the paper provide sufficient information on the computer resources (type of compute workers, memory, time of execution) needed to reproduce the experiments?
    \item[] Answer: \answerYes{}
    \item[] Justification: Yes, we clearly specify the computational resources used in our experiments. As indicated in Table \ref{table:time_comparison}, all experiments were conducted using a single NVIDIA A6000 GPU (1×A6000), providing sufficient information for reproducibility.

\item {\bf Code of ethics}
    \item[] Question: Does the research conducted in the paper conform, in every respect, with the NeurIPS Code of Ethics \url{https://neurips.cc/public/EthicsGuidelines}?
    \item[] Answer: \answerYes{}
    \item[] Justification: 
     Yes, this research fully conforms to the NeurIPS Code of Ethics. The work involves developing novel methods for 3D point cloud processing and does not involve human subjects, personal data collection, or sensitive applications that could cause harm. The code has been made available in an anonymized repository for reproducibility and transparency. The research methodology, experimental design, and reporting follow ethical standards for machine learning research.

\item {\bf Broader impacts}
    \item[] Question: Does the paper discuss both potential positive societal impacts and negative societal impacts of the work performed?
    \item[] Answer: \answerYes{}
    \item[] Justification: The paper discusses positive impacts in enabling quality control for personalized manufacturing and healthcare (Sections 1 and 5), does not explicitly address potential negative societal impacts. Future works may consider implications like potential job displacement in quality control or privacy concerns in personalized healthcare applications.

\item {\bf Safeguards}
    \item[] Question: Does the paper describe safeguards that have been put in place for responsible release of data or models that have a high risk for misuse (e.g., pretrained language models, image generators, or scraped datasets)?
    \item[] Answer: \answerNA{}
    \item[] Justification: The methods proposed in the paper focus on anomaly detection in manufacturing and do not involve high-risk models or datasets that could be misused. The untrained methods operate on 3D point cloud data of manufactured parts and do not pose risks associated with privacy, misinformation, or other harmful applications.

\item {\bf Licenses for existing assets}
    \item[] Question: Are the creators or original owners of assets (e.g., code, data, models), used in the paper, properly credited and are the license and terms of use explicitly mentioned and properly respected?
    \item[] Answer: \answerYes{}
    \item[] Justification: The paper cites original works for methods used in experiments (e.g., PointSGRADE, FPFH, BTF, CPMF) and includes proper references to these works in the bibliography. However, it does not explicitly mention licenses or terms of use for any datasets or code implementations that might have been utilized.

\item {\bf New assets}
    \item[] Question: Are new assets introduced in the paper well documented and is the documentation provided alongside the assets?
    \item[] Answer: \answerNA{}
    \item[] Justification: The paper primarily introduces theoretical frameworks and analyzes existing methods rather than releasing new datasets, code, or models as assets. The focus is on the mathematical formulation and experimental validation of untrained approaches for anomaly detection.

\item {\bf Crowdsourcing and research with human subjects}
    \item[] Question: For crowdsourcing experiments and research with human subjects, does the paper include the full text of instructions given to participants and screenshots, if applicable, as well as details about compensation (if any)? 
    \item[] Answer: \answerNA{}
    \item[] Justification: The research does not involve crowdsourcing or human subjects. It focuses on algorithmic approaches for anomaly detection in 3D point cloud data through untrained machine learning methods.

\item {\bf Institutional review board (IRB) approvals or equivalent for research with human subjects}
    \item[] Question: Does the paper describe potential risks incurred by study participants, whether such risks were disclosed to the subjects, and whether Institutional Review Board (IRB) approvals (or an equivalent approval/review based on the requirements of your country or institution) were obtained?
    \item[] Answer: \answerNA{}
    \item[] Justification: The research does not involve human subjects or clinical trials, so IRB approval was not required. The work is purely computational and focuses on machine learning methods for anomaly detection.

\item {\bf Declaration of LLM usage}
    \item[] Question: Does the paper describe the usage of LLMs if it is an important, original, or non-standard component of the core methods in this research? Note that if the LLM is used only for writing, editing, or formatting purposes and does not impact the core methodology, scientific rigorousness, or originality of the research, declaration is not required.
    \item[] Answer: \answerNA{}
    \item[] Justification: The research does not utilize large language models (LLMs) as components of the core methods. The proposed approaches are based on statistical learning, geometric priors, and optimization techniques rather than language models or generative AI.

\end{enumerate}

\end{document}